# Addressing Design Issues in Medical Expert System for Low Back Pain Management: Knowledge Representation, Inference Mechanism, and Conflict Resolution Using Bayesian Network

Debarpita Santra[a], Jyotsna Kumar Mandal[a], Swapan Kumar Basu[b], Subrata Goswami[c]

[a]Department of Computer Science and Engineering, Faculty of Engineering, Technology and Management, University of Kalyani, Block C, Nadia, Kalyani, West Bengal, PIN - 741245, India

[b]Department of Computer Science, Institute of Science, Banaras Hindu University, Varanasi-221005, Uttar Pradesh, India

[c]ESI Institute of Pain Management, ESI Hospital Sealdah premises, 301/3 Acharya Prafulla Chandra Road, Kolkata – 700009, West Bengal, India

**Abstract.** Aiming at developing a medical expert system for low back pain management, the paper proposes an efficient knowledge representation scheme using frame data structures, and also derives a reliable resolution logic through Bayesian Network. When a patient comes to the intended expert system for diagnosis, the proposed inference engine outputs a number of probable diseases in sorted order, with each disease being associated with a numeric measure to indicate its possibility of occurrence. When two or more diseases in the list have the same or closer possibility of occurrence, Bayesian Network is used for conflict resolution. The proposed scheme has been validated with cases of empirically selected thirty patients. Considering the expected value 0.75 as level of acceptance, the proposed system offers the diagnostic inference with the standard deviation of 0.029. The computational value of Chi-Squared test has been obtained as 11.08 with 12 degree of freedom, implying that the derived results from the designed system conform the homogeneity with the expected outcomes. Prior to any clinical investigations on the selected low back pain patients, the accuracy level (average) of 73.89% has been achieved by the proposed system, which is quite close to the expected clinical accuracy level of 75%.

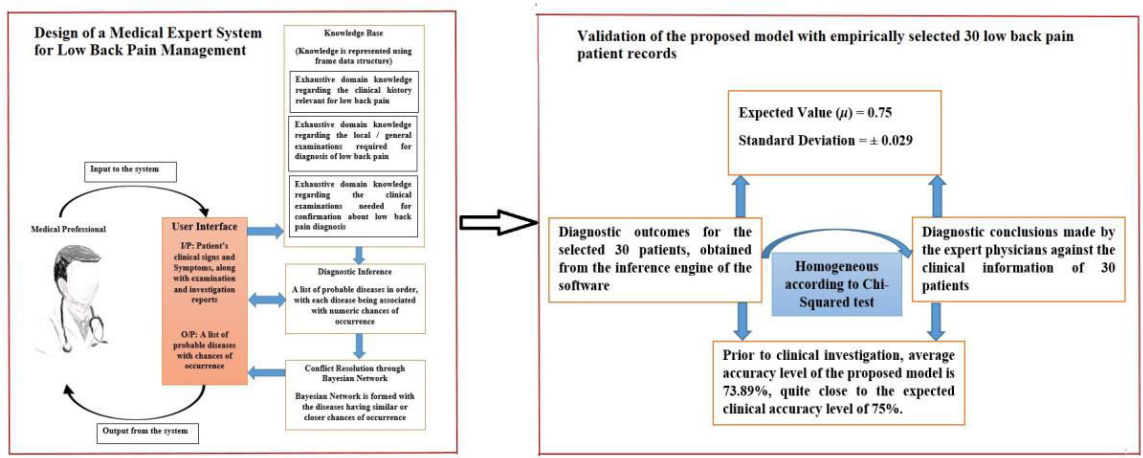

Figure: Graphical abstract for the methodology proposed in the manuscript

## 1. Introduction

Low Back Pain (LBP) [1] is a global health hazard that deprives many individuals of living their normal and routine lives [2]. With LBP being the major contributor to India's burden of disabling conditions [3], people like farmers, mill workers, porters, labourers, blacksmiths, goldsmiths, zari workers who are mostly from the lower socioeconomic strata of the society, suffer badly from this disease. LBP is also hastily affecting the affluent civilization in India. Treatment of LBP is challenging as it demands extremely specialized and updated knowledge about the intricate anatomical and physiological structure of a human body [4]. The issue can be resolved to a greater extent by using a medical expert system [5] using Artificial Intelligence (AI) [6] for performing reliable diagnosis with justified therapy recommendations for LBP diseases. This kind of knowledge-intensive software would be able to assist the general physicians, junior doctors, nurses and other care-givers in many primary and secondary healthcare settings in India, by providing quick and reliable medical consultancy service to deliver effective therapies for LBP.

Application of AI in the LBP domain is still in embryonic stage. With an aim to develop a reliable and affordable medical expert system for LBP management (MES-LBP) for the Indian milieu, this paper primarily deals with the design issues of how the acquired knowledge can be represented efficiently ensuring easy retrieval of knowledge, and how the retrieved knowledge can be efficiently processed by the system to produce reliable diagnostic conclusions. MES-LBP shall contain four building blocks: *user interface* (UI), *working memory* (WM), *knowledge base* (KB), and *inference engine* (IE) [7]. When an LBP patient comes to a physician for medical consultation, the attending physician would input relevant clinical information about the patient to the software through UI. The clinical information collected through UI is stored in the computational storage module WM. Previous case records of the patients are also stored for future reference. KB stores exhaustive and up-to-date medical knowledge for diagnosing LBP, gathered mostly from expert physicians and the existing literature. IE mines the huge volume of relevant knowledge from KB, matches the patient information with them and infers evidence-based diagnostic decisions. The diagnostic decisions, which are also warehoused, made visible to the attending physician through UI. Figure 1 shows the block diagram for MES-LBP.

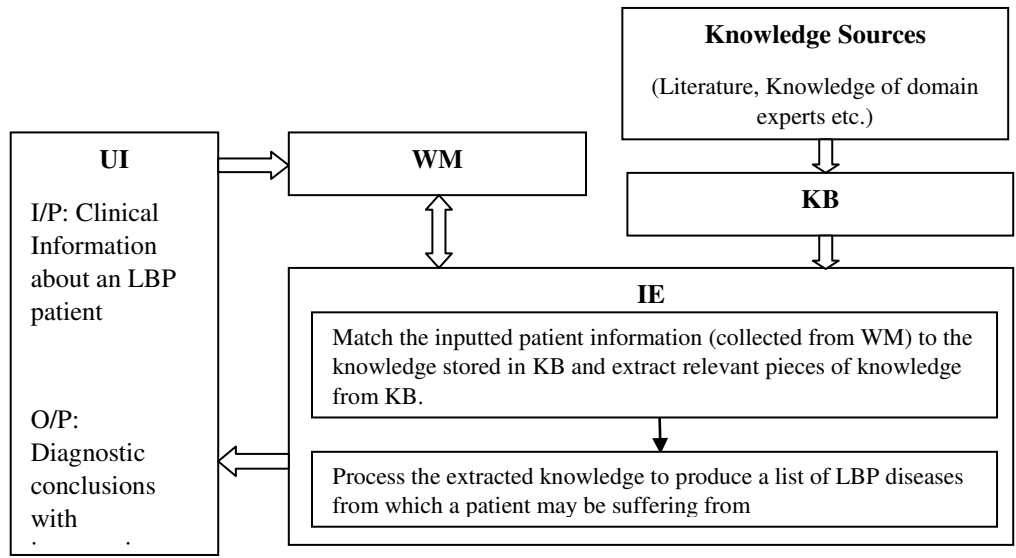

Fig 1: Block diagram of MES-LBP

Achieving reliability in MES-LBP requires proper design of each building block. With no standard guideline for LBP diagnosis being followed in India as well as other developing countries, the intended software follows a generalized procedure, as depicted in figure 2, for treating LBP patients. This procedure, followed in some pain clinics in India, has a clinically approved efficacy in LBP diagnosis.

In this paper, a knowledge representation scheme has been proposed for storing the acquired knowledge in KB of MES-LBP. An efficient reasoning strategy has been derived to be followed by IE. IE outputs a list of probable diseases, with each disease being associated with a numeric measure for chance of occurrence for the patient. As many LBP diseases may contain similar symptoms, the output disease list may also have a number of diseases with similar or closer chance of occurrence. These kinds of conflict resolution are done through IE using Bayesian Network.

This paper is structured as follows: section 2 provides a background of the work, section 3 discusses about the design issues of KB, section 4 proposes an inference methodology, section 5 gives the results and discussion, and section 6 concludes the paper.

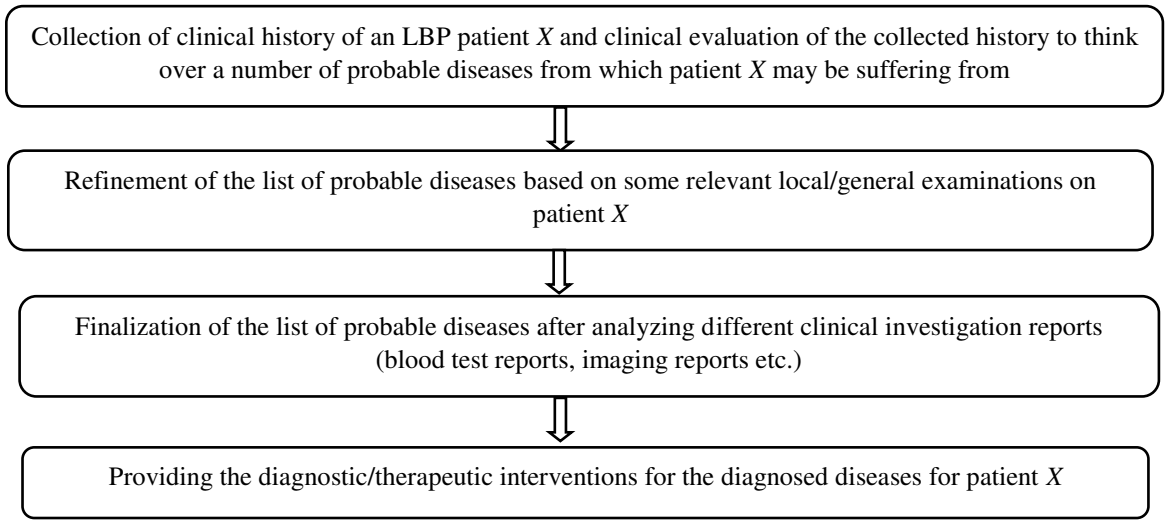

Fig 2: Generalized methodology for diagnosing LBP diseases

## 2. Background Study

MYCIN [8] is an early expert system and is one of the well-known expert systems to aid physicians in antimicrobial drugs for treatment of blood infections. The Causal Associational NETwork (CASNET) [9] is a computer system designed for diagnosis of glaucoma. INTERNIST [10] is a consultation programme for use in internal medicine. ONCOCIN [11], developed in Stanford's School of Medicine, was intended for use by oncologists and was used at Stanford's Oncology Clinic for protocol management for patients already diagnosed with cancer and being treated with chemotherapy. PUFF [12] is a small expert system for pulmonary functions, which comprises about 400 rules as well as 75 medical parameters. A medical expert system also exists in the literature to assist dermatologists on assessment of some of the skin diseases namely Psoriasis, Ichthyosis, Eczema, Meningitis etc. [13] and there is tele-monitoring system suitable for use in mobile phones to monitor the conditions that may lead to heart failure [14]. Coming to the LBP domain, an integrated methodology was proposed for an expert system to be used in managing LBP in future [15], and an expert system was designed to successfully diagnose intensity of LBP [16].

In recent years, there are some notable advancements where AI has been used in solving different classification and prediction issues of medical diagnosis. A hash-table based efficient and intelligent diagnosis algorithm for coronary artery disease has been proposed to distinguish between the healthy and diseased classes, and to predict the chances of occurrence of this disease among males and females [17]. A dynamic neural network has been developed for accurately estimating skeletal muscle forces, which can be incorporated in a decision support system for supporting functional analysis of muscle forces in real-time [18]. A logistic regression based prediction model has been designed as a smart risk assessment tool to deal with Down syndrome and neural tube anomalies [19]. A hybrid learning model was proposed for precise prediction of sleep staging, while the accuracy of the model has been achieved higher compared to the traditional classification approaches [20].

Development of an effective knowledge base poses a great challenge to the expert-system researchers. Many knowledge representation techniques are available in the literature. Among them, production rules are extensively used in medical domain and resemble as a good starting point for real-life understanding of the expert systems [21]. Since 1970s, rule-based knowledge representation is being used in many medical expert systems such as MYCIN, PUFF, CMDS [22], and so on [23]. Sometimes, production rules have been associated with some probabilistic measures to deal with uncertainties in medical knowledge.

Inference mechanisms used by the well-known medial expert systems are mainly rule based, semantic network based, and neural network based. Rule-based systems like MYCIN, CASNET, INTERNIST etc. use "if-then-else" constructs as constituents of production rules and take help of forward or backward chaining mechanisms for reaching to the conclusions. This kind of inferencing technique is very efficient and suitable when the domain knowledge is complete, consistent and unambiguous. In case of inconsistency or incomplete domain knowledge, the inferencing technique may not be able to reach to definitive conclusions. To overcome this problem, some expert systems use probabilistic inferencing mechanism using Bayesian Network [24]. A Bayesian Network, which follows the structure of a semantic network in terms of a directed acyclic graph, can be applied in medical domain for capturing the probabilistic relationships between clinical parameters (symptoms, test reports etc.) and diseases in a directed acyclic graph. Estimating the conditional probabilities of different nodes in such a Bayesian Network is very important and challenging. These a-priori probabilities are calculated from the disease profiles and patient's clinical records found in different hospitals. In this case, the considered data set should be unbiased and free of noise for achieving great accuracy in clinical decision making. But obtaining this kind of dataset is practically infeasible, as medical domain is full of uncertainties. Also, a clear understanding between the clinical parameters and diseases is very much needed as the Bayesian Network is formed with the causal components. This kind of problem can also be overcome by the use of neural network [25] as an inferencing mechanism. Neural network does not demand a clear understanding about the relationships between the input and output variables. Neural network learns from a huge amount of training data set and outputs satisfactorily. Many clinical decision support systems today use this inferencing mechanism. The larger the training data set is, the more accurate is the outcome of the inference. The shortcoming of a neural network is that it does not follow any particular execution logic and is not explicitly comprehensible. Also, it may not be possible to get a huge dataset for training the medical expert system.

For LBP, it is hard to generate a huge dataset from the hospitals in the third world countries for training a medical expert system. In the third world country like India, people mainly from the lower socio-economic strata of society do not opt for diagnosis of their LBP in the hospitals. LBP being a neglected medical domain from the part of stake holders, most of the hospitals does not maintain the records of LBP patients. So, due to lack of systematic clinical datasets for LBP in these countries, use of neural network as the inferencing logic is not a good choice. Rather, a hybrid inferencing mechanism using the rule-based technique combined with Bayesian approach can be a better solution in this context.

## 3. Proposed Knowledge Representation Scheme

Development of MES-LBP using rule-based methodology is the prime objective. Relevant medical knowledge for the expert system for LBP management is acquired from the existing literature (journals/articles/international guidelines) and expert physicians from the domain of LBP through finite clinical attributes $n$ ($>0$). The attributes are identified through repeated consultations with domain experts. Examples of some clinical attributes are: 'history of trauma', 'site of pain', 'type of pain', 'duration of pain', 'pain worsening factors', 'pain relieving factors', and so on. Each attribute can hold one or more than one value simultaneously. For example, the attribute 'history of trauma' will have only one value between 'yes' and 'no' at a particular time depending on the application scenario. Another attribute 'site of pain' may hold more than one value at a certain instant of time, as pain can be felt at many primary sites such as lumbar region, buttock, and greater trochanter [26].

For easy analogical reasoning, the frame data structure [27] has been used for representing the acquired knowledge. With the consideration of $n$ clinical attributes, a class frame named "$C_D$-frame" is defined with $n$ slots corresponding to $n$ attributes, and their respective values. A slot may have one or more than one value. As the $C_D$-frame holds exhaustive clinical information in terms of clinical attributes relevant for LBP domain, and the all-possible clinically accepted values for the attributes, it acts as a universal set $U_{lbp}$ for the LBP domain.

Mathematically, the $C_D$-frame can be expressed using 3-tuple $<A, V, F>$, where

- $A = \{a_i | a_i$ is a clinical attribute with $1 \leq i \leq n\}$ is a set of slot variables
- $V = \{v_{ij} | v_{ij}$ is the $j$-th value of attribute $a_i\}$, and $A \cap V = \phi$
- $F$ is a slot function defined as: $F \subseteq (A \times V)$ such that $\forall a_i \in A$ ($1 \leq i \leq n$), there is at-least one element in $V$.

It is assumed that there is a finite set of $x$ ($>0$) LBP diseases in the literature. Information about any disease $d_k$ ($1 \leq k \leq x$) is also represented in another class frame $D_k$ derived from the $C_D$-frame. With information about every disease being stored in individual derived class frame, each derived frame inherits all the slot variables and disease-specific attribute values from the $C_D$-frame. So, each derived class frame is basically a sub-set of $U_{lbp}$.

Mathematically, it can be said that $U_{lbp} = \cup_{k=1 \text{ to } x} D_k$, where $D_k$ ($\subset U_{lbp}$) corresponds to the derived class frame for the disease $d_k$. A slot variable $a_i$ ($1 \leq i \leq n$) in the derived frame $D_k$ may accept zero, one, or greater than one value. If the set of slot variables in $D_k$ is denoted as $A_k$, and the set of slot values as $V_k$, where $V_k = \{v_{il} | v_{il}$ is the $l$-th value of attribute $a_i$ in $D_k\}$, the following logical derivations can easily be made:

i) $\cap_{k=1 \text{ to } x} A_k = A$ and $\cup_{k=1 \text{ to } x} A_k = A$
ii) $V = \cup_{k=1 \text{ to } x} V_k$, and

iii) $\cap_{k=1 \text{ to } y} V_k \neq \phi$ where $y (\leq x)$ represents the no. of non-empty slot variables in $D_k$, as many LBP diseases may have common slot values

If the slot variable $a_i$ in $D_k$ has greater than one value, all the values may not have the same clinical weightage. Also, every slot variable may not carry the same clinical significance for diagnosing different LBP diseases. For example, consider three attributes relevant for LBP diagnosis: 'site of pain' ($a_1$), 'pain worsening factors' ($a_2$), and 'bowel/bladder habit' ($a_3$). For an LBP disease 'Sacroiliac Joint Arthropathy' [28] (SIJA) ($d_1$), the clinical significance of the $a_1$ is high, followed by $a_2$ and $a_3$. While the attribute $a_1$ for SIJA may have values as 'buttock region' ($v_{11}$), 'low back area' ($v_{12}$), and 'leg' ($v_{13}$), it is clinically evidenced that most of the patients with SIJA feel pain in the buttock region, with less number of patients complaining pain confined in low back area, and very rare patients having pain in legs. So, if weightages are to be assigned to these slot values, the value $v_{11}$ will gain highest weightage, followed by the slot values $v_{12}$ and $v_{13}$. Let us consider the attribute $a_2$ for SIJA, which would have values as 'lying on the affected side' ($v_{21}$), 'sitting for greater than 15 minutes' ($v_{22}$), and 'supine position' ($v_{23}$) arranged according to their weightages in descending order. The third attribute $a_3$ will have only one value 'normal' ($v_{31}$) at a particular instant of time; therefore, it gets full clinical weightage. The clinical significance ($c_s$) of slot variables and the assignment of clinical weightage ($c_w$) to the slot values vary disease-wise. Both the clinical significance and the clinical weightage are quantified through natural numbers, starting from 1, which indicates the lowest priority. The disease $d_1$ can be represented as

$$d_1 := [<a_1, \{(v_{11}, c_w^{11} = 3), (v_{12}, c_w^{12} = 2), (v_{13}, c_w^{13} = 1)\}, c_s^1 = 3>, <a_2, \{(v_{21}, c_w^{21} = 3), (v_{22}, c_w^{22} = 2), (v_{23}, c_w^{23} = 1)\}, c_s^2 = 2>, < a_3, \{(v_{31}, c_w^{24} =1)\}, c_s^3 = 1>] \quad (1)$$

From Eqn. (1), it can be conceptualized that, besides inheriting the slot variables and the relevant attribute values from the $C_D$-frame, the derived class frame $D_k$ for disease $d_k$ extends the base frame by incorporating the clinical weightage with each slot-value, and the clinical significance for each slot-variable.

Formally, the derived class frame $D_k$ is represented using 6-tuple $<A_k, CS_k, \gamma, V_k, CW_k, \delta>$, where

- $A_k (\subseteq A)$ is the set of slot variables in $D_k$ and $|A_k| = n$.
- $CS_k \subset N$, where $N$ consists of only natural numbers, assigned as clinical significance to the slot variables in $A_k$; $\forall c_s^i \in CS_k$ with $1 \leq i \leq n$, $1 \leq c_s^i \leq |A_k|$ if the respective slot value field is non-empty, otherwise $c_s^i = 0$.
- $\gamma$ is a mapping function from $A_k$ to $CS_k$ such that $\forall a_i \in A_k$ ($1 \leq i \leq n$), there is exactly one element from $CS_k$.
- $V_k (\subseteq V)$ is the set of attribute-values in $D_k$, and is defined as $V_k = \{v_{il} | v_{il}$ is the $l$-th $(0 \leq l \leq m)$ value of attribute $a_i$ in $D_k\}$ and $|V_k| = m$.
- $CW_k \subset N$, assigned as clinical weightage to the slot values in $V_k$, and is defined as $CW_k = \{c_w^{il} | c_w^{il}$ is the clinical weightage assigned to the $l$-th $(0 \leq l \leq m)$ value of attribute $a_i$ $(1 \leq i \leq n)$ in $D_k\}$. $\forall c_w^{il} \in CW_k$, $1 \leq c_w^{il} \leq m$ if the respective slot value field is non-empty, otherwise $c_w^{il} = 0$.
- $\delta$ is a mapping function from $V_k$ to $CW_k$ such that $\forall v_{il} \in V_k$ where $1 \leq i \leq n$ and $1 \leq l \leq m$, there is exactly one element from $CW_k$.

This is to be noted that $A_k \cap V_k = \phi$, and $CS_k \neq CW_k$. This kind of representation of a disease actually resembles a number of atomic production rules related to the disease. As the disease $d_k$ is characterized by $n$ clinical attributes, the $j$-th $(1 \leq j \leq n)$ tuple of the frame $D_k$ with the slot variable $a_j$, clinical significance $c_s^j$, $y^{\diamond}$ non-empty slot values associated with clinical weightages is basically a production rule ($r_j$) of the form shown in (2).

$$r_j: \langle a_j, \prod_{p=1 \text{ to } P}(v_{jp}, w_{jp}), c_s^j \rangle \rightarrow d_k \tag{2}$$

The production rule system corresponding to $D_k$ is represented as a set $P_D^k = \{(C_j, d_k) | C_j \ (1 \leq i \leq n)$ represents the antecedent part in (2)$\}$.

LBP diagnosis of a patient $X$ goes through mainly three sequential phases: *phase* 1 for collection and analysis of relevant clinical history (termed as *h*-phase), *phase* 2 for the study of local/general examination reports (termed as *g*-phase), and *phase* 3 for analysis of clinical investigation reports (termed as *i*-phase). KB of the intended medical expert system should hold the knowledge about each phase. Knowledge about the *h*-phase states about the correlation between different LBP diseases and the clinical history parameters. For example, if the site of pain of a patient is buttock, then the patient may be suffering from SIJA [28]. Here, 'site of pain' is a clinical history parameter/attribute. Knowledge about the *g*-phase depicts the correspondence between the general examination parameters and the LBP diseases. For example, if the crossed SLR test [29] report of an LBP patient is positive, then the chance of the patient having Prolapsed Inter-vertebral disc disease (PIVD) [30] is high. Here, 'crossed SLR test report' is a general examination parameter. Knowledge about the *i*-phase finds association between the pathological investigation reports and the LBP diseases. For example, if the HLA-B27 test [31] report of a young patient is positive, then he/she may be suffering from Ankylosing Spondylitis [32]. Here, 'HLA-B27 test report' is a pathological examination parameter. So, for each phase, there are some dedicated clinical attributes. More specifically, if $A$ is the set of $n$ clinical attributes, a subset $A_1$ ($\subset A$) of $n_1$ ($> 0$) attributes would be constructed for capturing knowledge about the *h*-phase, a subset $A_2$ ($\subset A$) of $n_2$ ($> 0$) attributes would be constructed for capturing knowledge about the *g*-phase, and another subset $A_3$ ($\subset A$) of $n_3$ ($= n - (n_1 + n_2)$) attributes would be formed for capturing knowledge about the *i*-phase. Here, $A_1 \cap A_2 \cap A_3 = \phi$, and $A_1 \cup A_2 \cup A_3 = A$. With $x$ ($>0$) LBP diseases in the literature, there would be three different sub-frames for each disease, designed using the derived class frames like $D_k$, for representing the knowledge about the three different phases. These frames are termed as sub-class frames. The sub-class frame associated with the *h*-phase is denoted as $f_h$, which holds $n_1$ slot variables; the sub-class frame related with the *g*-phase is denoted as $f_g$, which holds $n_2$ slot variables; and the third sub-class frame for representing the knowledge about *i*-phase is denoted as $f_i$, which holds $n_3$ slot variables. There would be at-most $x$ instantiations for each of the three sub-class frames. Formally, the *h*-frame, *g*-frame, or *i*-frame for a disease $d_k$ following the same 6-tuple representation as $D_k$ is represented as $\langle A^q_k, CS^q_k, \gamma^q, V^q_k, CW^q_k, \delta^q \rangle$, where $q \in \{n_1, n_2, n_3\}$, and $|A^q_k| = q$, $|CS^q_k| = q$, $|V^q_k| = \sum_{i=1 \text{ to } q} |\tilde{V}_i|$ ($\tilde{V}_i$ represents the set of values for the *i*-th slot), and $|CW^q_k| = q$.

KB holds a linked structure of the frames to maintain integriety. The starting point of the linked structure is a frame named "root frame" ($f_{root}$), which contains only three slots in the names of three phases. Value of each slot of $f_{root}$ holds a pointer to another frame called "sub-root frame" ($f_{sub-root}$). There are basically three sub-root frames: $f^h_{sub-root}$ for *phase* 1, $f^g_{sub-root}$ for *phase* 2, and $f^i_{sub-root}$ for *phase* 3. The $f^h_{sub-root}$ frame holds $x$ slots $\{d_j \mid d_j\ (1 \leq j \leq x)$ is an LBP disease$\}$, with value of each slot pointing to the corresponding instance frame ($f_I$) belonging to the set of $x$ instances of the *h*-frame $\{f^j_{I-h} | f^j_{I-h}\ (1 \leq j \leq x)$ is the *j*-th instance frame of $f^h\}$. The $f^g_{sub-root}$ frame also holds $x$ slots for $x$ diseases, with each slot value pointing to the respective instance frame belonging to the set of $x$ instances of the *g*-frame $\{f^j_{I-g} | f^j_{I-g}\ (1 \leq j \leq x)$ is the *j*-th instance frame of $f^g$ corresponding to disease $d_j\}$. Similarly, the $f^i_{sub-root}$ frame contains $x$ slots corresponding to $x$ diseases, and each slot value points to the corresponding instance frame belonging to the set of $x$ instances of the *i*-frame $\{f^j_{I-i} | f^j_{I-i}\ (1 \leq j \leq x)$ is the *j*-th instance frame of $f^i\}$. With an $f^j_{I-z}$ frame ($z \in \{h, g, i\}$, and $1 \leq j \leq x$) being formally represented similar to a sub-class frame ($f_z$), the mathematical definitions of $f_{root}$ and $f^g_{sub-root}$ are given as follows.

The "root frame" $f_{root}$ is defined using 3-tuple $\langle S_R, V_R, F_R \rangle$, where

- $S_R = \{s^i_r | s^i_r \ (1 \leq i \leq 3)$ represents the slot variable corresponding to the $i$-th phase$\}$
- $V_R = \{\&(f^z_{sub-root}) | \&(f^z_{sub-root}) \ (z \in \{h, g, i\})$ represents the address of the $f^z_{sub-root}$ frame$\}$
- $F_R$ is a mapping function between $S_R$ and $V_R$ such that $\forall s^i_r \in S_R \ (1 \leq i \leq 3)$, there is exactly one element from $V_R$.

Similar to $f_{root}$, the $f^z_{sub-root}$ ($z \in \{h, g, i\}$) is mathematically defined using 3-tuple $<S_S, V_S, F_S>$, where
- $S_S = \{s^i_s | s^i_s \ (1 \leq i \leq x)$ represents the slot variable corresponding to disease $d_i\}$
- $V_S = \{\&(f^j_{I-z}) | \&(f^j_{I-z})$ represents the address of $f^j_{I-z}$ corresponding to disease $d_i\}$
- $F_S$ is a mapping function between $S_S$ and $V_S$ such that $\forall s^i_s \in S_S \ (1 \leq i \leq x)$, there is exactly one element from $V_S$.

As a whole, KB can be formally represented as a set of 8-tuple $<f_{root}, \hat{S}_f, \hat{I}_f, \hat{E}, K_b, \check{N}, \hat{U}, \ddot{I}_R>$, where

- $f_{root}$ is the root frame from which the search to KB starts.

- $\hat{S}_f = \{f^z_{sub-root} | f^z_{sub-root} \ [z \in \{h, g, i\}]$ is either $f^h_{sub-root}$, or $f^g_{sub-root}$, or $f^i_{sub-root}\}$ is the set of all sub-root frames.

- $\hat{I}_f$ is the set of instance frames and is denoted as $\hat{I}_f = I_h \cup I_g \cup I_i$, where $I_h = \{f^j_{I-h} | f^j_{I-h} \ (1 \leq j \leq x)$ is the $j$-th instance frame of $f^h\}$, $I_g = \{f^j_{I-g} | f^j_{I-g} \ (1 \leq j \leq x)$ is the $j$-th instance frame of $f^g\}$, and $I_i = \{f^j_{I-i} | f^j_{I-i} \ (1 \leq j \leq x)$ is the $j$-th instance frame of $f^i\}$.

- $\hat{E}$ is the set of edges to make the inter-connections inside the frame systems composed of $f_{root}$, $\hat{S}_f$, and $\hat{I}_f$. There are a total of $(3x+3)$ links in the whole structure, as 3 links are coming out from $f_{root}$ to point $f^h_{sub-root}$, $f^g_{sub-root}$, and $f^i_{sub-root}$; $x$ links are coming out from each sub-root frame to point $x$ instance frames.

- $K_b$ is starting point for reasoning with the knowledge stored in KB.

- $\check{N}$ is the set of all slot variables in KB and is denoted as $\check{N} = \{K_b\} \cup Var_1 \cup Var_2$, where $Var_1 = S_R \cup S_S$ and $|Var_1| = (3x+3)$, and $Var_2 = \cup_{k=1 \text{ to } x} A_k = A$, where $A_k = \cup_{q=1 \text{ to } 3} A^q_k$

- $\hat{U}$ is the set of slot values for all the instance frames and is denoted as $\hat{U} = V_h \cup V_g \cup V_i$, where $V_h = \{Val^k(f^j_{I-h}) | Val^k(f^j_{I-h})$ is the set of values at the $k$-th slot of the $j$-th instance of the $h$-frame$\}$, $V_g = \{Val^k(f^j_{I-g}) | Val^k(f^j_{I-g})$ is the set of values at the $k$-th slot of the $j$-th instance of the $g$-frame$\}$, $V_i = \{Val^k(f^j_{I-i}) | Val^k(f^j_{I-i})$ is the set of values at the $k$-th slot of the $j$-th instance of the $i$-frame$\}$.

- $\ddot{I}_R$ is a set of reasoning rules of the form $\alpha \rightarrow \beta$, where $\alpha \in \check{N}$, and $\beta \in (\check{N} \cup \hat{U})^*$. The reasoning rules define the strategy how KB is searched starting from $K_b$. The rules are as follows:
    i) $K_b \rightarrow \wedge_{i=1 \text{ to } 3} s^i_r$ \quad // $s^i_r \in S_R \ [1 \leq i \leq 3]$ is the $i$-th slot variable in $f_{root}$ and '$\wedge$' denotes the logical operator 'AND'
    ii) $s^i_r \rightarrow \wedge_{j=1 \text{ to } x} s^j_s$ \quad // $s^j_s \in S_s \ [1 \leq j \leq x]$ is the $j$-th slot variable in $f^z_{sub-root}$, where $z$ represents the $i$-th phase
    iii) $s^j_s \rightarrow \wedge_{k=1 \text{ to } q} (a^j_k, c^j_{s k})$ \quad // $a^j_k \in A^q_j \ [q \in \{n_1, n_2, n_3\}]$ is the $k$-th $[1 \leq k \leq q]$ clinical attribute in $f^j_{I-z}$ and $c^j_{s k}$ is the clinical significance of the considered attribute

iv)          $a^j_k \rightarrow \wedge_{l=1 \text{ to } m} (val_l, c^l_w)$ // $val_l \in Val^k(f^j_{I-z})$ is the $l$-th value of attribute $a^j_k$ and $c^l_w$ is the clinical weightage of $val_l$ and $|Val^k(f^j_{I-z})| = m$

In each reasoning rule except the fourth one, the variables at the right hand side should be explored one by one starting from the left. More specifically, the left-most variable in the consequent part will be explored first, then the second, and so on.

## 4. Inference Mechanism

This section focuses on deriving an efficient match logic for IE and also on proposing execution methodology of the matched knowledge to provide reliable diagnostic conclusions. As output, IE shows the names of the diseases with their chance of occurrence in decreasing order. The chance of occurrence of every disease is a numeric measure lying between 0 and 1 both inclusive, and is obtained from the result of the matching process. If there is a full match, i.e. the patient information completely matches with the knowledge about a disease in KB, then the chance of occurrence for that disease for the particular patient is 1. If the patient information partially matches with the stored knowledge about a disease, then the chance of occurrence for the disease would be greater than 0 but less than 1. On the contrary, if, for another disease, there is no match between the patient information and the stored knowledge about the disease, the chance of occurrence for that disease is 0.

### 4.1 Matching Process for Inferencing

The matching process can be visualized as a black box, where the patient information is provided as the input, and the output is the matched knowledge that are retrieved from KB. The matching process is done phase-wise; that is, at the first phase, an LBP patient's clinical history is taken into account, whereas, the second phase collects the information related to the general/local examinations performed on the patient, and during the third phase, only the patient's investigation reports are collected, if any.

Information about a patient will be collected through the clinical parameters belonging to set $A$. The inputted patient information is also structurally represented using frames [33]. The frame that stores patient information is named as "$I/P$-frame", and three categories of $I/P$ frames are designed in this context: $I/P$-frame$_h$ that captures patient's clinical history through the clinical attributes in $A_1$, $I/P$-frame$_g$ that captures patient's local/general examination information using the attributes mentioned in $A_2$, $I/P$-frame$_i$ that stores patient's clinical investigation reports through the attributes mentioned in $A_3$. There is an obvious possibility that some of the clinical parameters may not be relevant for different LBP patients. In this case, the respective slot values of the frames will be empty. The frames $I/P$-frame$_h$, $I/P$-frame$_g$, and $I/P$-frame$_i$ contain all the slot variables that are present in the $h$-frame, $g$-frame, and $i$-frame respectively. The value fields corresponding to the slot variables in the input frames are supplied externally from the patients through UI.

An $I/P$-frame$_z$ ($z \in \{h, g, i\}$) is formally represented as 3-tuple $<A^q_z, \Gamma^q_z, \beta_z>$,

- $A^q_z (\subseteq A)$ is the set of slot variables relevant for phase $z$, where $q \in \{n_1, n_2, n_3\}$
- $\Gamma^q_z$ is the set of slot values for the attributes in $A^q_z$, and is defined as:
  $\Gamma^q_z = \{inp_{ij} | inp_{ij}$ is the $j$-th value inputted against attribute $a_i$ in $A^q_z\}$, and $A^q_z \cap \Gamma^q_z = \phi$
- $\beta_z$ is a slot function defined as: $\beta_z \subseteq (A^q_z \times \Gamma^q_z)$ such that $\forall a_i \in A^q_z$ ($1 \leq i \leq |A^q_z|$), there is at-least one element in $\Gamma^q_z$.

When there exists an I/P-frame$_h$ corresponding to an LBP patient X, IE starts matching with all the instances of h-frame ($I_h$) stored in KB. More specifically, first all the slots of $f^1_{I-h}$ are visited and the information regarding how much information in the instance frame matched with the patient's information is noted. After visiting all slots of $f^1_{I-h}$, visit to the slots of $f^2_{I-h}$ is started, thus the match information is noted. Proceeding in this way, all the slots of the $f^x_{I-h}$ are visited, and the corresponding match information is noted. The entire match information is stored in a 2d matrix called $M_h$ with $n_1$ rows and x columns. Each row corresponds to a slot variable / clinical attribute, and each column corresponds to an LBP disease. $M_h$ matrix will be different for every LBP patient. The entry $M_h(i,j)$ for patient X has two parts: the first part denoted as $M_h(i,j).value$ holds the i-th slot-value(s) of $f^j_{I-h}$ corresponding to disease $d_j$ that have matched against the i-th slot value(s) at I/P-frame$_h$ for patient X, and the second part denoted as $M_h(i,j).cs$ holds the clinical significance $r^j_i$ of the i-th attribute $a_i$ for disease $d_j$.

If the i-th slot variable of $f^j_{I-h}$ for disease $d_j$ holds a set of values $Val^i(f^j_{I-h})$ and the i-th slot variable of I/P-frame$_h$ for patient X holds a set of values $V_X$, then the entry $M_h(i,j)$ for patient X would hold a set of values $V_{result} = V_X \cap Val^i(f^j_{I-h})$, along with the clinical significance $r^j_i$ of $a_i$ for $d_j$. Similar to the $M_h$ matrix, there will be an $M_g$ matrix and a $M_i$ matrix of size ($n_2 \times x$) and ($n_3 \times x$) respectively, as there are $n_2$ attributes in $f^j_{I-g} \in I_g$ ($1 \leq j \leq x$) and $n_3$ attributes in $f^j_{I-i} \in I_i$ ($1 \leq j \leq x$). The algorithm for matching an I/P-frame$_h$ with $I_h$ is given in *Algorithm* 1.

**Procedure Matching :** *Algorithm* 1

*Input*: I/P-frame$_h$ which holds patient X's clinical history, and $I_h$ which is the set of x instances of h-frame
*Output*: The matrix $M_h$ for patient X, which holds all the match information disease-wise
*Method*: The algorithm matched information and structurally kept the matched knowledge in matrix $M_h$
       for their easy retrieval
**Begin**
    For j=1 to x do    // x is the total number of LBP diseases in the literature
        For i=1 to $n_1$ do    // $n_1$ is the fixed number of attributes in $f^j_{I-h}$
           If $Val^i(f^j_{I-h}) \neq \phi$ then    // $Val^i(f^j_{I-h})$ is the set of values of i-th slot at $f^j_{I-h}$
              $V_{result} = V_X \cap Val^i(f^j_{I-h})$ //$V_X$ is the set of values corresponding to the i-th slot at I/P-frame$_h$
              $M_h(i,j).value = V_{result}$ //$M_h(i,j).value$ indicates the first part of the entry $M_h(i,j)$
              $M_h(i,j).cs = r^j_i$    // $M_h(i,j).cs$ indicates the second part of the entry $M_h(i,j)$
           End If
        End For
    End for
    Return $M_h$
**End Matching**

The time complexity of *Algorithm* 1 is $O(x*n_1)$. The following section shows how the knowledge stored in $M_h$, $M_g$, or $M_i$ are executed to determine the chance of occurrence of each disease.

## 4.2 Execution of Matched Knowledge

The matched knowledge in $M_h$ would be accessed column wise. For each column where at-least one entry with value other than $\phi$ is found, a numerical measure called 'match strength' (*ms*) is obtained, where $0 < ms \leq 1$. Depending on the value of *ms*, it is determined how much match has been occurred. To describe the implication of the value of *ms*, a linguistic variable '*match*' is used, which accepts only three linguistic values '*full*', '*partial*', and '*zero*'.

$$match = \begin{cases} \text{`full' match with disease } d_i, & \text{if } ms == 1 \text{ for column } i\ (1 \leq i \leq x) \text{ in } M_h \\ \text{`partial' match with disease } d_i, & \text{if } ms == f\ (0 > f < 1) \text{ for column } i\ (1 \leq i \leq x) \text{ in } M_h \\ \text{`zero' match with disease } d_k, & \text{if } ms == 0 \text{ for column } i\ (1 \leq i \leq x) \text{ in } M_h \end{cases}$$

An algorithm for calculating the match strength for each column in $M_h$ is given in *Algorithm* 2.

**Procedure Match_Strength:** *Algorithm* 2

*Input*: Matrix $M_h$ holding the matched knowledge against the inputted clinical history for patient X.
*Output*: A list $L_h$ of probable diseases along with their chances of occurrence for patient X.
*Method*: This algorithm basically finds out how much match occurs there for each entry in $M_h$. The $M_h(i,j).cs$ for every *i*-th row and *j*-th column of the $M_h$ matrix is taken into account for calculating the match strength. A column of the $M_h$ matrix with at-least one non-empty entry signifies probable presence of a disease and is included in the list $L_h$. $L_h$ is the list of diseases obtained at the end of *phase* 1 after the execution on matched knowledge is performed.

**Begin**
  *Count* = 0// *Count* is a temporary variable
  For *j* =1 to *x* do  // *x* is the total no. of rows in $M_h$
    For *i* =1 to $n_1$ do  // $n_1$ is the total no. of columns in $M_h$
      If $M_h(i,j) \neq \phi$ then
        *Count* ++;
        *Temp* (*Count*) = *j*     //*Temp* is a temporary 1-d array with number of elements less or equal to *x*
        *Break* ();
      End If
    End For
  End For

  $L_h = \phi$   // $L_h$ is the list of probable diseases diagnosed from clinical history
  For *k* = 1 to *Count* do
    *index* = *Temp*(*k*)  // *index* is a temporary variable whose value is used to retrieve the appropriate instance of *h*-frame among all the instances in $I_h$
    $Frame_{temp} = f^{index}_{I-h}$  // $Frame_{temp}$ is a temporary frame of similar structure of the *h*-frame
    $ts_k = 0$   // $ts_k$ is a temporary variable
    $ls_k = 0$   // $ls_k$ is a temporary variable
    For *l*=1 to $n_1$ do
      $max_l$ = Maximum($c_w^{l1}$, $c_w^{l2}$,…, $c_w^{lv}$) //$max_l$ is the maximum of all the weightages associated with the values of *l*-th slot of $Frame_{temp}$, and *v* is the total number of slot values
      $ts_k = ts_k + (max_l * r_l)$  //$r_l$ denotes the clinical significance of attribute $a_l$
      If $M_h(k,l) \neq \phi$ then
        $imax_l$ = Maximum(w)   //$imax_l$ is the maximum of all the weightages associated with the values of $M_h(k,l)$ where w $\subseteq \{c_w^{l1}, c_w^{l2},…, c_w^{lv}\}$
      $ls_k = ls_k + (imax_l * M_h(k,l).cs)$
      End If

```
    End For
    ms_k = ls_k / ts_k       // ms_k is the match strength of disease d_k
    If  ms_k ≠ 0 then
        L_h = L_h ∪ {(d_k, ms_k)}
    End If
  End For
  Return L_h
End Match_Strength
```

The time complexity of *Algorithm* 2 is $O(x*n_1)$. In the same way, the match strength algorithms for the matrices $M_g$ and $M_i$ are executed to obtain the list of probable diseases $L_g$ and $L_i$ respectively.

So, individually and independently the lists of probable diseases are obtained for each phase. Finally, only one resultant list $L_D$ is produced from all the lists using *Algorithm* 3. During the design of the algorithm, it has been kept in mind that a patient is diagnosed with a disease only after the clinical investigation. Before the clinical investigations but after the clinical examinations, the phycisians become around 75% sure about the chance of occurrence of a disease. Before the clinical examinations, the physicians give 50% assurance about the chance of occurrence of a disease. In this context, the diseases found in $L_i$ get highest priority (priority value: 3), followed by the diseases in $L_g$ (priority value: 2), and the diseases in $L_h$ (priority value: 1). In ideal case, if a disease specification fully matches ($ms = 1$) with the inputted patient information, then the total priority value would be $[(1* 1) + (1*2) + (1*3)] = 6$.

**Procedure Provisional_Diagnosis:** *Algorithm 3*

*Input*: $L_h$, $L_g$, $L_i$ which are the lists of probable diseases obtained from *phase* 1, *phase* 2, and *phase* 3 respectively

*Output*: $L_D$ which is the final list of probable diseases

*Method*: The chance of occurrence for a disease in $L_h$ is multiplied by 1, whereas the chance of occurrence for that disease in $L_g$ is multiplied by 2, and the chance of occurrence for that disease in $L_i$ is multiplied by 3 to get its measure of chance of occurrence for $L_D$. Finally, numeric value obtained against the disease in $L_D$ is divided by 6 to get the actual result, provided all the three phases are considered.

**Begin**

```
 D = L_h.d ∩ L_g.d ∩ L_i.d    //L_h.d, L_g.d, L_i.d represent any disease present in L_h, L_g, or L_i respectively
 T_D = φ                       // T_D is an empty set initially
 For i = 1 to |D| do           // |D| represents the cardinality of set D
    c_h = 1 * L_h.D_i.ms       //L_h.D_i.ms represents the match strength of disease D_i belonging to D
    c_g = 2 * L_g.D_i.ms       //L_g.D_i.ms represents the match strength of disease D_i belonging to D
    c_i = 3 * L_i.D_i.ms       //L_i.D_i.ms represents the match strength of disease D_i belonging to D
    c = c_h + c_g + c_i        // c is a temporary variable
    T_D = T_D ∪ {(D_i, c)}
 End For
 L_h = L_h − D
 L_g = L_g − D
 For i = 1 to |L_g| do
    L_g.d_i.ms = 2 * L_g.d_i.ms   // d_i represents the i-th disease in L_g
```

```
  End For
L_i = L_i − D
 For j = 1 to |L_i| do
    L_i.d_j.ms = 3 * L_i.d_j.ms   // d_j represents the j-th disease in L_i
 End For
T_D = T_D ∪ L_h ∪ L_g ∪ L_i
L_D = ϕ
For k = 1 to |T_D| do
  disease = T_D[k].d    // disease is a temporary variable
  chance = float(T_D[k].cs / 6)  // chance is a temporary variable
  L_D = L_D ∪ {(disease, chance)}
End For
Return L_D
End Provisional_Diagnosis
```

The set $L_D$ may contain more than one disease with same or closer match strengths. In this case, the Bayesian Network is used to resolve the conflicts.

### 4.3 Conflict Resolution using Bayesian Network

Bayesian Network (BN) is key computing technology for dealing with uncertainties in AI [34]. A BN is a directed acyclic graph, where the nodes symbolize a set of random variables $X = X_1, X_2, \ldots, X_n$ from the application domain, and the directed arcs join pairs of nodes $X_i \rightarrow X_j$ $(i \neq j)$, signifying the direct dependencies between the corresponding random variables. Two nodes should maintain a direct link in-between if one affects or causes the other, with the arc specifying the direction of the effect. Assuming discrete variables, the strength of the relationship between the variables is computed by conditional probability distributions of each node.

The joint probability distributions of the nodes in BN is denoted as $p(X_1 = x_1, X_2 = x_2, \ldots, X_n = x_n)$, or $p(x_1, x_2, \ldots, x_n)$. The chain rule of probability theory is applied to factorize joint probabilities using Eqn. (3).

$$p(X_1 = x_1, X_2 = x_2, \ldots, X_n = x_n) = p(x_1) * p(x_2|x_1) * \ldots * p(x_n|x_1, \ldots, x_{n-1}) \quad (3)$$
$$= \prod_{i=1 \text{ to } n} p(x_i | x_1, \ldots, x_{i-1})$$
$$= \prod_{i=1 \text{ to } n} p(x_i | Parents(x_i)), \text{ where } Parents(x_i) \text{ is a subset of } \{x_1, \ldots, x_{i-1}\}$$

For the application domain, a BN is constructed using *Algorithm* 4 involving those diseases in the set $L_D$ for which conflicts have been found.

**Procedure Construction_BN:** *Algorithm* 4

*Input*: The input frames $I/P\text{-frame}_h$, $I/P\text{-frame}_g$, and $I/P\text{-frame}_i$; the sets of clinical attributes $A_1$, $A_2$, and $A_3$; the set Đ of $y$ LBP diseases for which conflicts have been found in $L_D$.
*Output*: A Bayesian Network $BN_{LBP}$ for the considered application domain
*Method*: The algorithm constructs a BN in an incremental way, first by allocating the nodes corresponding to the conflicted diseases, and then by assigning conditional probability tables with each node and making casual connections among the nodes. The probabilities in the conditional probability tables are obtained from the domain knowledge.

**Begin**
   $BN_{LBP} = \phi$     // The BN does not contain any node initially
   $N_D = \phi$     // The set $N_D$, which is initially empty, may contain the BN nodes corresponding to $y$ diseases
   For $i = 1$ to $y$ do
      $N^i_D \leftarrow getnode()$ // The function $getnode()$ is used to create a node in; $N^i_D$ is the $i$-th element in $N_D$
      $N_D = N_D \cup \{N^i_D\}$
   End For
   $BN_{LBP} = BN_{LBP} \cup N_D$
   $N_H = \phi$     // The set $N_H$, which is initially empty, would hold the nodes corresponding to $n_1$ clinical history parameters belonging to $A_1$
   For $i_1 = 1$ to $n_1'$ do     // $n_1'$ denotes the no. of non-empty slots in $I/P$-frame$_h$ and $n_1' \leq n_1$
      $N^{i1}_H \leftarrow getnode()$ // $N^{i1}_H$ is the $i_1$-th element in $N_H$
      $N_H = N_H \cup \{N^{i1}_H\}$
   End For
   $BN_{LBP} = BN_{LBP} \cup N_H$
   $N_G = \phi$     // The set $N_G$, which is initially empty, may hold the nodes corresponding to $n_2$ parameters corresponding to the patient's general/local examination reports belonging to $A_2$
   For $i_2 = 1$ to $n_2'$ do     // $n_2'$ denotes the no. of non-empty slots in $I/P$-frame$_g$ and $n_2' \leq n_2$
      $N^{i2}_G \leftarrow getnode()$ // $N^{i2}_G$ is the $i_2$-th element in $N_G$
      $N_G = N_G \cup \{N^{i2}_G\}$
   End For
   $BN_{LBP} = BN_{LBP} \cup N_G$
   $N_I = \phi$     // The set $N_I$, which is initially empty, holds the nodes corresponding to $n_3$ parameters corresponding to the patient's clinical investigation reports belonging to $A_3$
   For $i_3 = 1$ to $n_3'$ do     // $n_3'$ denotes the no. of non-empty slots in $I/P$-frame$_i$ and $n_3' \leq n_3$
      $N^{i3}_I \leftarrow getnode()$ // $N^{i3}_I$ is the $i_3$-th element in $N_I$
      $N_I = N_I \cup \{N^{i3}_I\}$
   End For
   $BN_{LBP} = BN_{LBP} \cup N_I$

   For $i = 1$ to $|N_H|$ do
      $cpt(N^i_H) = p(a_i)$     // $cpt(N^i_H)$ denotes the conditional probability table associated with node $N^i_H$ and $p(a_i)$ denotes the probability of attribute $a_i \in A_1$
      $Parents(N^i_H) \leftarrow \phi$   // $Parents(N^i_H)$ denotes the parents of node $N^i_H$
   End For
   For $j = 1$ to $|N_D|$ do
      $Parents(N^j_D) \leftarrow N_H$   //$Parents(N^j_D)$ denotes the parents of node $N^j_D$
      $cpt(N^j_D) = p(d_j | A_1)$   // $cpt(N^j_D)$ denotes the conditional probability table associated with node $N^j_D$ and $p(d_j | A_1)$ denotes the conditional probability for disease $d_j \in Đ$
   End For
   For $k = 1$ to $|N_G|$ do
      $Parents(N^k_G) \leftarrow N_D$   // $Parents(N^k_G)$ denotes the parents of node $N^k_G$
      $cpt(N^k_G) = p(a_k | Đ)$   // $cpt(N^k_G)$ denotes the conditional probability table associated with node $N^k_G$ and $p(a_k | Đ)$ denotes the conditional probability for clinical attribute $a_k \in A_2$
   End For
   For $l = 1$ to $|N_I|$ do
      $Parents(N^l_I) \leftarrow N_D$   // $Parents(N^l_I)$ denotes the parents of node $N^l_I$
      $cpt(N^l_I) = p(a_l | Đ)$   // $cpt(N^l_I)$ denotes the conditional probability table associated with node $N^l_I$ and

$p(a_l | Đ)$ denotes the conditional probability for clinical attribute $a_l \in A_3$

   End For
Return $BN_{LBP}$
End **Construction_BN**

The complexity of *Algorithm* 4 is $O(b)$, where $b = \text{Maximum}(n_1', n_2', n_3', y)$. The joint probabilities for each node belonging to $N_D$ are determined for the resolution of conflicts. For a disease $d_i$ ($1 \leq i \leq y$), the corresponding joint probability is calculated as shown in Eqn. (4). For ease of understandability, the $i$-th ($1 \leq i \leq q$, $q \in \{n_1, n_2, n_3\}$) element of the set $A_1$, $A_2$, and $A_3$ is denoted as $A_1^i$, $A_2^i$, and $A_3^i$ respectively.

$p(A_1^1, A_1^2, \ldots, A_1^{n1}, d_i, A_2^1, A_2^2, \ldots, A_2^{n2}, A_3^1, A_3^2, \ldots, A_3^{n3})$

$= \prod_{i=1 \text{ to } n1} p(A_1^i) * p(d_i | A_1^1, A_1^2, \ldots, A_1^{n1}) * \prod_{i=1 \text{ to } n2} p(A_2^i | d_i) * \prod_{i=1 \text{ to } n3} p(A_3^i | d_i)$      (4)

The joint probability values for the conflicted diseases are compared to each other, and the disease with largest joint probability value is considered to have the highest chance of occurrence for the patient, and the disease with smallest joint probability value has the lowest chance of occurrence.

A simple instance from LBP domain is taken here for illustrating the mechanism for reliable inference through BN. For the sake of simplicity, five clinical attributes for *phase* 1, and three attributes for *phase* 2 have been considered here, without taking into account any attribute for *phase* 3. The five clinical attributes under consideration for *phase* 1 are: 'site of pain' ($a_1$), 'type of pain' ($a_2$), 'pain referred zone' ($a_3$), 'pain radiation zone' ($a_4$), and 'pain at rest' ($a_5$). The three attributes for *phase* 2 are: 'SLR test report' ($a_6$), 'FABER test report' ($a_7$), and 'FADIR test report' ($a_8$) [35-36]. In this example, there is consideration of only four LBP diseases: 'Facet Joint Arthropathy' [37] ($d_1$), 'Discogenic Pain' ($d_2$) [38], 'Sacroiliac Joint Arthropathy' ($d_3$), and 'Prolapsed Intervertebral Disc with Radicular Pain' ($d_4$). The knowledge about these four diseases has been represented using the proposed scheme. After execution of *Algorithm* 1, *Algorithm* 2, and *Algorithm* 3 on the stored knowledge against the clinical information of an LBP patient, the final list $L_D$ of probable diseases is assumed to contain the diseases $d_1$, $d_2$, $d_3$, and $d_4$ with the match strengths being calculated as 0.94, 0.94, 0.70, and 0.53 respectively. It is easily visible from $L_D$ that the match strengths for both $d_1$ and $d_2$ are same. So, there is a conflict between the diseases $d_1$ and $d_2$. This kind of conflict is resolved using the BN as shown in figure 3. The conditional probability tables in figure 3 have been constructed using the domain knowledge.

From figure 3, the joint probabilities for $d_1$ and $d_2$ are calculated as follows.

$p(a_6 = \text{'normal'}) \& (a_7 = \text{'negative'}) \& (a_8 = \text{'negative'}) \& (d_1 = \text{'true'}) \& (a_1 = \text{'low back'}) \& (a_2 = \text{'dull'}) \& (a_2 = \text{'aching'}) \& (a_3 = \text{'buttock'}) \& (a_3 = \text{'posterior thigh'}) \& (a_5 = \text{'no'})) \approx 0.09$      (5)

$p((a_6 = \text{'normal'}) \& (a_7 = \text{'negative'}) \& (a_8 = \text{'negative'}) \& (d_2 = \text{'true'}) \& (a_1 = \text{'low back'}) \& (a_2 = \text{'dull'}) \& (a_2 = \text{'aching'}) \& (a_3 = \text{'buttock'}) \& (a_3 = \text{'posterior thigh'}) \& (a_5 = \text{'no'})) \approx 0.12$      (6)

So, it can be easily concluded from (5) and (6) that though the match strengths of both the diseases are same, the occurrence probability for disease $d_2$ is high compared to that of $d_1$.

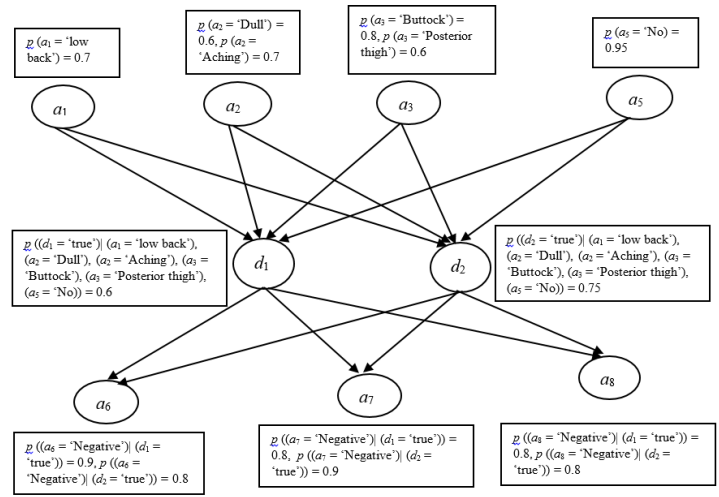

Fig 3: Construction of BN for conflict resolution among the diseases $d_1$ and $d_2$

## 5. Results and Discussion

The proposed methodology for the design of IE for MES-LBP has been tested with clinical information of thirty LBP patients collected from the ESI Hospital Sealdah, Kolkata – 700009, West Bengal, India. The study group comprised only those LBP patients who have been diagnosed with SIJA, FJA, PIVD, DP, or Piriformis Syndrome (PS) [39]. There are patients who have been diagnosed with multiple LBP diseases simultaneously. The disease distribution among the considered patients is shown in figure 4.

The patient records have been accessed with prior ethical approval from the hospital authority. Fifteen important clinical attributes have been considered for capturing clinical history of patients, and fourteen important attributes are considered for acquiring information related to the clinical inspections/examinations performed on the patients. The investigation (blood/imaging) part of diagnosis has not been considered here.

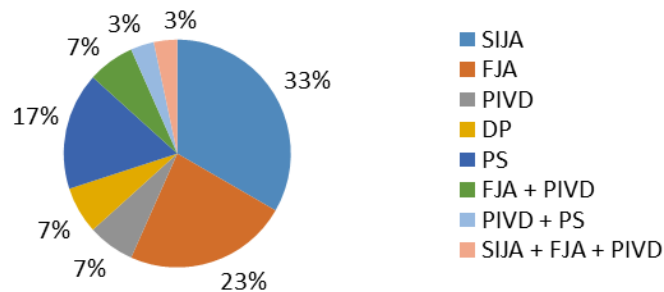

Fig 4: The disease distribution among the thirty patients under consideration

Provisional diagnosis for the considered LBP patients has been performed by a number of certified pain specialists at the same hospital. In most of the cases, the treating physicians concluded with only one disease which has the highest possibility of occurrence. But in reality, an LBP patient may suffer from multiple diseases, among which some diseases may have very low chance of occurrence. Neglecting those diseases which are at their early stages may not be a good practice; rather taking some preventive means would reduce the chance of further developing the diseases for the patients. The proposed inferencing mechanism has been implemented, and validated using the thirty patient cases. The software outcome for each of the thirty patient cases includes a number of possible diseases with their respective non-conflicting chances of occurrence. For each patient, the top five diagnostic outcomes of the implemented modules have been considered. The diagnostic inference of the software has been compared with the conclusions made by the expert physicians.

With the assumption that the expert physicians diagnose with equal to or more than 75% of accuracy prior to clinical investigations, the expected value ($\mu$) of concluding that a patient suffering from an LBP disease is 0.75. The implication is that an expert physician would diagnose an LBP patient with disease $d$ only if the chance of occurrence of $d$ for the patient is equal or greater than 0.75. With the set of thirty LBP patients $\{P_k | P_k \ (1 \leq k \leq 30)$ is an LBP patient under consideration$\}$, the patient-wise distribution of diseases diagnosed by the expert physicians as well as by the software along with their respective chances of occurrence is shown in figures 5(a), 5(b), and 5(c).

Denoting the software outcomes as observed results, it should be measured how much the observed results deviate from the expected value $\mu$. This kind of deviation is called standard deviation (*SD*) [40], and it is calculated using Eqn. (7).

$$SD = \frac{\sqrt{\Sigma(|x - \mu|)^2}}{N} \qquad (7)$$

where, $x$ is a value in the data set, and $N$ is the total no. of data points in the considered population of 30 patients.

While calculating the *SD*, the observed results have been compared against the expert outcomes. Suppose, the outcomes by expert physicians and the outcomes by the software have been kept in sets $O_E$ and $O_S$ respectively which are defined as follows:

$O_E = \{(d^e, ch^e) | \ d^e$ is an LBP disease diagnosed by expert physicians and $ch^e \ (0.75 \leq ch^e \leq 1.0)$ is the chance of occurrence of $d^e\}$

$O_S = \{(d^s, ch^s) | \ d^s$ is an LBP disease diagnosed by the software and $ch^s \ (0 \leq ch^s \leq 1.0)$ is the chance of occurrence of $d^s\}$

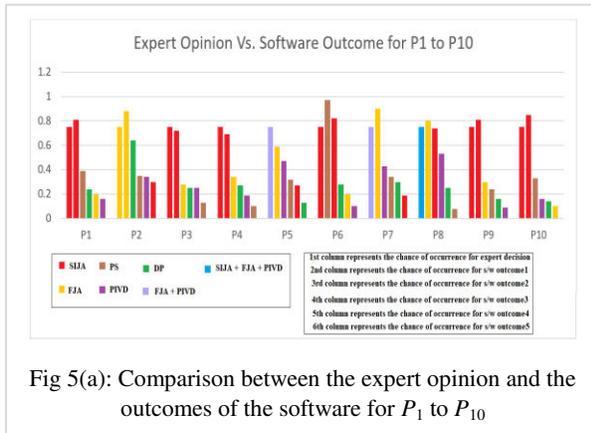

Fig 5(a): Comparison between the expert opinion and the outcomes of the software for $P_1$ to $P_{10}$

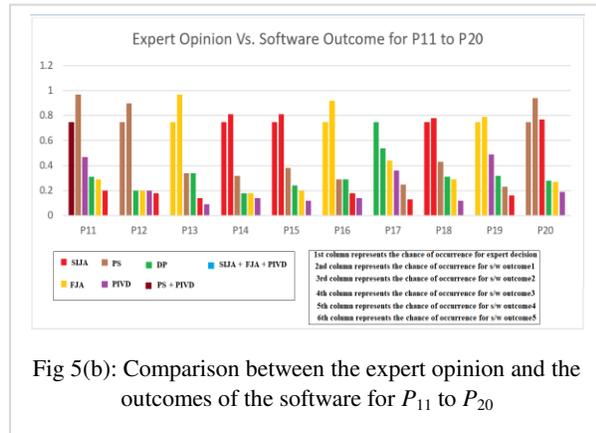

Fig 5(b): Comparison between the expert opinion and the outcomes of the software for $P_{11}$ to $P_{20}$

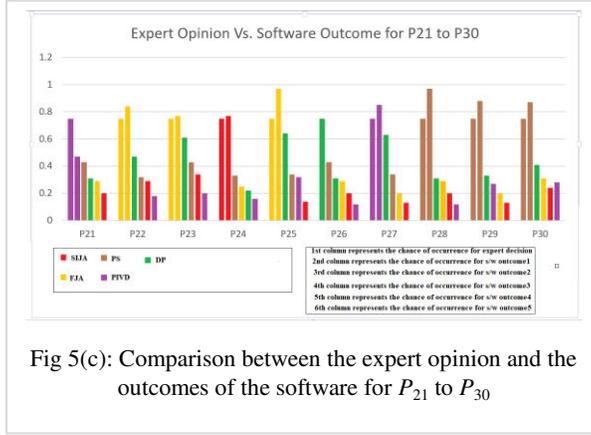

Fig 5(c): Comparison between the expert opinion and the outcomes of the software for $P_{21}$ to $P_{30}$

If a patient is diagnosed with a disease $d$ by expert physicians, it is observed whether $d \in O_S$. If yes, then the chance of occurrence as measured by the software is compared with $\mu$. If $d \notin O_S$, then it is assumed that the observed chance of occurrence for $d$ is 0. With four patients ($P_5$, $P_7$, $P_8$, and $P_{11}$) being diagnosed by more than one disease by the expert physicians, there is a total of 35 diagnostic results for 30 patients. So, the total no. of data points ($N$) is 35. Using Eqn. (7), the $SD$ for the considered scenario is calculated as 0.029. The patient-wise deviation of the observed chances of occurrence from $\mu$ is shown in figure 6.

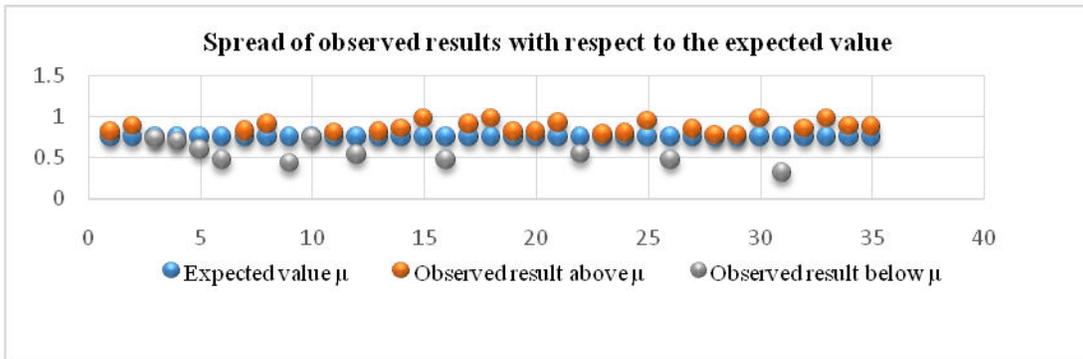

Fig 6: The deviation of observed results with respect to $\mu = 0.75$

It is clearly visible from figure 6 that among the 35 data points, 24 observed results reside above the level of expected value, while only 11 observed results reside below the clinically acceptable threshold of 0.75.

Now, the Pearsonian Chi-square test [41] for homogeneity is performed to see whether the pattern of the observed results as shown in figure 6 are alike to expert opinions or expected outcomes. As part of the test, it has been first observed how the distribution of diseases varied among the thirty patients age-group wise. The parameters 'less than 20 years of age', 'between 20 to 40 years of age', 'between 40 to 60 years of age', and 'above 60 years of age' have been denoted by $v_1^0$, $v_1^1$, $v_1^2$, and $v_1^3$ respectively.

Table 1 shows the disease distribution among the considered patients diagnosed by both the expert physicians and the software.

Table 1: Distribution of LBP diseases age-group wise

| Name of the disease | No. of patients diagnosed by expert physicians | | | | | No. of patients diagnosed by the software | | | | |
|---|---|---|---|---|---|---|---|---|---|---|
| | $v_1^0$ | $v_1^1$ | $v_1^2$ | $v_1^3$ | Total | $v_1^0$ | $v_1^1$ | $v_1^2$ | $v_1^3$ | Total |
| SIJA | 0 | 4 | 6 | 1 | **11** | 0 | 4 | 6 | 0 | **10** |
| FJA | 0 | 1 | 8 | 1 | **10** | 0 | 1 | 8 | 1 | **10** |
| PIVD | 0 | 2 | 4 | 0 | **6** | 0 | 2 | 4 | 0 | **6** |
| DP | 0 | 0 | 2 | 0 | **2** | 0 | 0 | 1 | 0 | **1** |
| PS | 0 | 0 | 5 | 1 | **6** | 0 | 0 | 6 | 2 | **8** |
| **Total** | **0** | **7** | **25** | **3** | **35** | **0** | **7** | **25** | **3** | **35** |

For each cell in the second column named "No. of patients diagnosed by expert physicians" in table 1, the expected frequency is calculated using Eqn. (8) and the calculations have been shown in table 2.

$$e_{ij} = \frac{o_i \times o_j}{N} \tag{8}$$

Where, $e_{ij}$ denotes the expected frequency, $o_i$ represents the marginal column frequency, $o_j$ signifies the marginal row frequency, and $N$ is the entire sample size.

Table 2: Expected frequency distribution of LBP diseases age-group wise

| Name of the disease | $v_1^0$ (approx.) | $v_1^1$ (approx.) | $v_1^2$ (approx.) | $v_1^3$ (approx.) | Total (approx.) |
|---|---|---|---|---|---|
| SIJA | 0 | 2.2 | 7.85 | 0.94 | **11.0** |
| FJA | 0 | 2 | 7.14 | 0.86 | **10.0** |
| PIVD | 0 | 1.2 | 4.29 | 0.51 | **6.0** |
| DP | 0 | 0.4 | 1.43 | 0.17 | **2.0** |
| PS | 0 | 1.2 | 4.29 | 0.52 | **6.0** |
| **Total** | **0** | **7.0** | **25.0** | **3.0** | **35.0** |

From tables 1 and 2, the Chi-Squared Static ($\chi^2$), which is calculated using Eqn. (9), is obtained as 11.08 with 'degrees of freedom' (*df*) for the considered scenario being $(i - 1) \times (j - 1) = 12$, where *i* and *j* are the no. of rows and columns respectively in table 2.

$$\chi^2 = \frac{\sqrt{\Sigma(|o_{ij} - e_{ij}|)^2}}{e_{ij}} \tag{9}$$

where, $o_{ij}$ is the observed frequency provided in the (*i,j*) cell of column 3 in table 1, and $e_{ij}$ is the expected frequency shown in the (*i,j*) cell in table 2.

With the considered accuracy level of 75%, the theoretical upper bound for homogeneity (critical value) with the *df* as 12 is 14.845. It can be concluded easily that; the obtained result is homogeneous. In general, the threshold value or the level of uncertainty is kept at 0.05%. In this case, the upper value is 21.026. In this case also, the obtained result shows homogeneity.

The efficiency of MES-LBP has been analysed using three metrics: *recall*, *precision*, and *accuracy* [42]. In the present case, the *recall* rate is used to measure the false negatives, and the *precision* rate to emphasize the false positives. For example, an expert physician diagnoses an LBP patient with three LBP diseases, and MES-LBP reaches to the conclusions with two diseases. If the two diseases diagnosed by the software are among the expert outcomes (decisions by expert physicians), then the recall rate is said to be 66.66%, and the precision rate is said to be 100%. The accuracy of the theoretical models is measured using Eqn. (10). For the sake of simplicity and easy visibility, only the software outcomes with chances of occurrence $ch \geq 0.75$ have been taken into account for computing three performance metrics.

$$accuracy\ rate = 2 \times \frac{precision\ rate \times recall\ rate}{precision\ rate + recall\ rate} \tag{10}$$

As per design considerations, the average recall rate of the proposed model is calculated as 74.44%, the average precision rate is obtained as 76.67%, and the average accuracy rate is achieved as 73.89%. The obtained accuracy rate is closer to the expected accuracy level of 75%.

## 6. Conclusions

The advantage of developing a rule-based medical expert system is that, the diagnostic conclusions provided as the output by the system closely match with the expected outcomes, provided the acquired knowledge is correct, consistent, and complete. As reliability is a major concern in medical expert systems, the proposed knowledge representation and the inferencing techniques act as a firm basis for development of the intended expert system for LBP management. The experimental results demonstrate that the accuracy rate of the software outcomes is close to the expected value, and the observed outcomes are homogeneous with the expected results. As there are limited numbers of LBP diseases as per expert knowledge, the time and space complexity would not be of much concern. As per the design considerations, the proposed schemes can be easily extended for design of a full-fledged expert system on LBP management in future. There are mainly two limitations of this work: first, only 30 patient cases have been considered for validation of the proposed work, and second, no clinical investigation results have been taken into account, leaving a scope of around 25% clinical uncertainty. With this paper proposing the methodology for conflict resolution during the inferencing of the software, other kinds of uncertainties would be handled in future for making the software more reliable. The models would be tested and validated with large number of LBP patients and their clinical investigation reports would also be considered.

**Acknowledgement**

The authors are sincerely thankful to the director and other faculty members at the ESI Institute of Pain Management, ESI Hospital Sealdah, West Bengal, India for providing exhaustive domain knowledge. Also, the authors are very grateful to the hospital authority (ESI Hospital) and the members of ethics committee for supporting this research by allowing to access sufficient patient records.

**Ethical approval:** "All procedures performed in studies involving human participants were in accordance with the ethical standards of the institutional and/or national research committee (ESI Institute of Pain Management Institutional Ethics Committee (IEC)/ Institutional Review Board (IRB) + reference number: 011/2018-19) and with the 1964 Helsinki declaration and its later amendments or comparable ethical standards."
**Acknowledgement**

The authors are sincerely thankful to the director and other faculty members at the ESI Institute of Pain Management, ESI Hospital Sealdah, West Bengal, India for providing exhaustive domain knowledge. Also, the authors are very grateful to the hospital authority (ESI Hospital) and the members of ethics committee for supporting this research by allowing to access sufficient patient records.

**Ethical approval:** "All procedures performed in studies involving human participants were in accordance with the ethical standards of the institutional and/or national research committee (ESI Institute of Pain Management Institutional Ethics Committee (IEC)/ Institutional Review Board (IRB) + reference number: 011/2018-19) and with the 1964 Helsinki declaration and its later amendments or comparable ethical standards."